\documentclass[lettersize,journal]{IEEEtran}
\usepackage{amsmath,amsfonts}
\usepackage{algorithmic}
\usepackage{algorithm}
\usepackage{array}
\usepackage[caption=false,font=normalsize,labelfont=sf,textfont=sf]{subfig}
\usepackage{textcomp}
\usepackage{stfloats}
\usepackage{url}
\usepackage{verbatim}
\usepackage{graphicx}
\usepackage{cite}
\usepackage{multirow}
\usepackage{tabularx}

\usepackage{xcolor}

\usepackage{colortbl}

\begin{document}

\title{POCE: Pose-Controllable Expression Editing}

\author{Rongliang Wu, Yingchen Yu, Fangneng Zhan, Jiahui Zhang, Shengcai Liao,~\IEEEmembership{Senior Member,~IEEE}, Shijian Lu$^*$
\IEEEcompsocitemizethanks{
\IEEEcompsocthanksitem R. Wu, Y. Yu, J. Zhang, and S. Lu are with the School of Computer Science and Engineering, Nanyang Technological University, Singapore.
\IEEEcompsocthanksitem F. Zhan is with the Nanyang Technological University, Singapore and Max Planck Institute for Informatics, Germany.
\IEEEcompsocthanksitem S.Liao is with the Inception Institute of Artificial Intelligence, Abu Dhabi, United Arab Emirates.
}

\thanks{* indicates the corresponding author. Email: shijian.lu@ntu.edu.sg}
}

\markboth{IEEE TRANSACTIONS ON IMAGE PROCESSING}%
{Wu \MakeLowercase{\textit{et al.}}: POCE: Pose-Controllable Expression Editing}


\maketitle

\begin{abstract}
Facial expression editing has attracted increasing attention with the advance of deep neural networks in recent years. However, most existing methods suffer from compromised editing fidelity and limited usability as they either ignore pose variations (unrealistic editing) or require paired training data (not easy to collect) for pose controls. This paper presents POCE, an innovative pose-controllable expression editing network that can generate realistic facial expressions and head poses simultaneously with just unpaired training images. POCE achieves the more accessible and realistic pose-controllable expression editing by mapping face images into UV space, where facial expressions and head poses can be disentangled and edited separately. POCE has two novel designs. The first is self-supervised UV completion that allows to complete UV maps sampled under different head poses, which often suffer from self-occlusions and missing facial texture. 
The second is weakly-supervised UV editing that allows to generate new facial expressions with minimal modification of facial identity, where the synthesized expression could be controlled by either an expression label or directly transplanted from a reference UV map via feature transfer.
Extensive experiments show that POCE can learn from unpaired face images effectively, and the learned model can generate realistic and high-fidelity facial expressions under various new poses. 
\end{abstract}


\section{Introduction}
Facial expression editing aims to edit the expression of a face image without changing the face identity. 
Automated and realistic expression editing has attracted increasing interest due to its wide range of applications in photography, animation, etc. However, generating high-fidelity expressions is a challenging task as the human visual system is very sensitive to tiny changes in facial expressions~\cite{mori2012uncanny,zakharov2019few}. While considering concurrent variations in facial expressions and head poses as in practical situations, realistic and high-fidelity facial expression editing becomes even more challenging.

Automated facial expression editing has achieved quite impressive progress in recent years. One typical approach is pose-fixed editing that focuses on expression editing only without handling the head pose of the edited face image. Leveraging the recent development of generative adversarial networks (GANs)~\cite{goodfellow2014generative}, several studies~\cite{choi2018stargan,pumarola2018ganimation,wu2020cascade,ling2020toward} formulate pose-fixed expression editing as an unpaired image-to-image translation task and require just a single source image for inference. However, these methods suffer from \textit{limited realism} as head poses and facial expressions usually vary simultaneously by nature. Additionally, they require the edited face image to be frontal or almost frontal and cannot handle many face images that are under non-frontal poses~\cite{choi2018stargan,pumarola2018ganimation,wu2020cascade,ling2020toward}. 

Pose-controllable expression editing aims to edit facial expressions and head poses simultaneously with minimal modification of facial identity features.
It has been attracting increasing interest from both academia and industry since it is better aligned with natural expression changes. 
Most existing studies exploit 3D facial structures as extracted by 3D modeling~\cite{blanz1999morphable} or deep generative networks~\cite{goodfellow2014generative}, but suffer from two typical constraints. First, they require paired images (i.e., face images of the same person with different expressions and poses)~\cite{geng2018warp,zhang2020freenet} or video sequences~\cite{burkov2020neural,ha2019marionette,zakharov2019few,siarohin2019first} for training, which are not easy to collect in practice and accordingly limit the \textit{usability} of these methods greatly. Second, they usually condition on facial landmarks that inherently encode facial expressions, facial identity and head poses altogether~\cite{zakharov2019few,zhang2020freenet,ha2019marionette}. 
The editing of such highly entangled expressions and poses in landmarks tends to introduce undesired modification of face identity (e.g., face shape), and this degrades the editing \textit{flexibility} and editing \textit{quality} greatly. 

This paper presents a novel \textbf{PO}se-{\textbf{C}ontrollable} \textbf{E}xpression editing (POCE) network that can edit facial expressions and head poses simultaneously with just unpaired training images. 
Inspired by the idea of UV maps that project 3D texture to a 2D pose-invariant template with universal per-pixel alignment, POCE converts face images into UV maps and disentangles expression editing and pose generation elegantly. Given a face image, we fit a 3D face model to sample a facial UV map which allows to edit expressions in the UV space and render the edited UV to new poses accurately. 
POCE has two novel designs that enable pose-controllable expression editing. 
The first is self-supervised UV completion that allows to generate complete UV texture from face images of different poses with various self-occlusions. 
The second is weakly supervised UV editing that allows to generate realistic expressions with minimal modification of face identity, where the synthesized expression could be controlled by either an expression label or directly transplanted from a reference UV map via feature transfer.
Extensive experiments show that POCE can achieve realistic pose-controllable expression editing with just unpaired training data.

The contributions of this work are threefold. 
\textit{First}, we propose POCE, an innovative network that can edit facial expressions and head poses simultaneously with just unpaired training images. 
\textit{Second}, we introduce UV maps for pose-controllable facial expression editing. On top of that, we design novel UV completing and UV editing techniques which achieve UV completion of self-occluded face images and identity-preservative expression editing, respectively. 
\textit{Third}, extensive experiments show that POCE achieves superior pose-controllable expression editing quantitatively and qualitatively.

\section{Related Work}

\subsection{Facial Expression Editing}
Automated expression editing has been studied for years and most existing works can be grouped into 3D model based methods and generation based methods.

\smallskip
\noindent{\bf 3D Model based Methods: }
Classical expression editing methods model 3D face structures with 3D Morphable Models (3DMMs). For example, \cite{blanz1999morphable} presents the first public 3DMM, where linear model was created to represent face variations.  
\cite{vlasic2006face} introduces a multi-linear model to map one person’s performance to facial animations of another. \cite{thies2016face2face} designs Face2Face for expression tracking and re-targeting. 3DMMs can jointly model expressions and poses, but they tend to produce blurs due to the Gaussian assumption~\cite{geng20193d}. In addition, they require hard-to-collect 3D face scans or videos in training, which limits their usability greatly.

\begin{figure*}[t]
\begin{center}
\includegraphics[width=1.\linewidth]{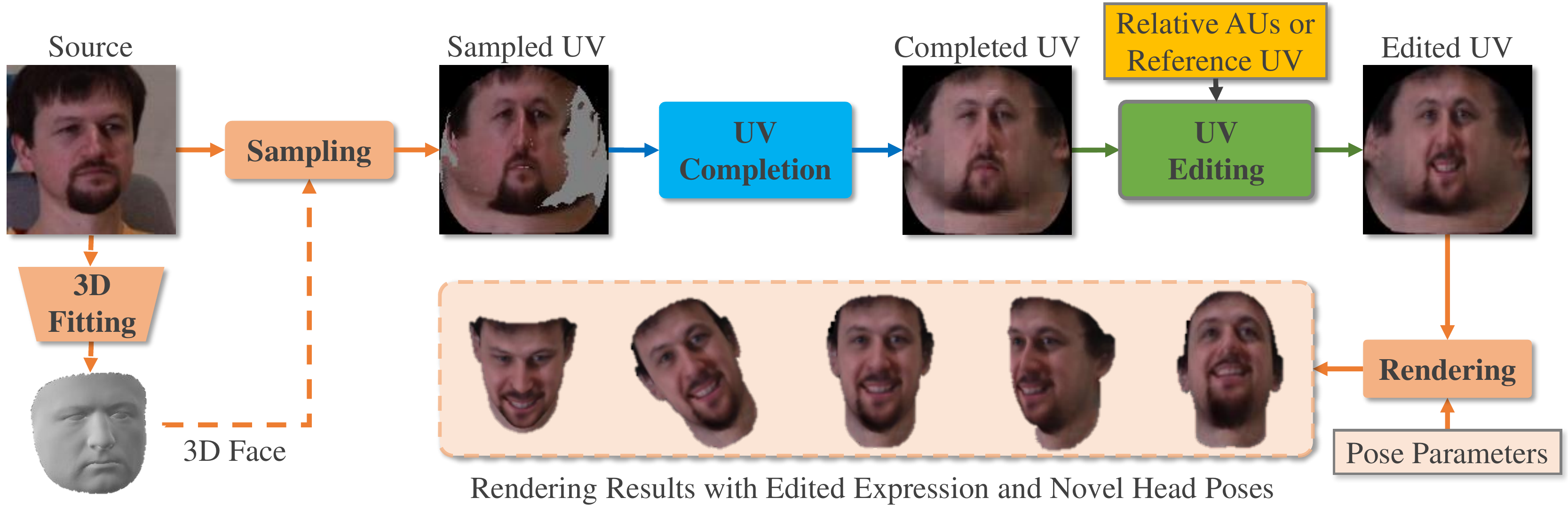}
\end{center}
\caption{
    The framework of the proposed POCE: Given a \textit{Source} face, we first fit a 3D face model to the face image to convert it to a \textit{Sampled UV} where expression editing and poses generation can be disentangled conveniently. The \textit{Sampled UV} is then fed to a \textit{UV Completion} module and a \textit{UV Editing} module which complete the UV map and edit the expression conditioned on \textit{Relative AUs} or transferred from \textit{Reference UV} via feature transfer, respectively. Finally, the \textit{Edited UV} is rendered to novel head poses conditioned on \textit{Pose Parameters} for generating realistic expressions and poses simultaneously.
}
\label{fig:pipeline}
\end{figure*}

\smallskip
\noindent{\bf Generation based Methods: }
Generation based methods exploit deep generative networks~\cite{goodfellow2014generative} for facial expression editing.  
For example, \cite{choi2018stargan} proposes StarGAN for multi-modality editing conditioned on discrete expression labels. 
\cite{pumarola2018ganimation} designs GANimation for continuous expression editing. 
\cite{wu2020cascade} presents Cascade EF-GAN for progressive expression editing. 
~\cite{wu2020leed} introduces LEED for label-free expression editing. 
\cite{ling2020toward} proposes MSF for fine-grained expression editing. These prior studies can work with unpaired images, but they can only handle pose-fixed editing which degrades the editing realism greatly as expressions and poses usually vary concurrently by nature. 

Pose-controllable editing via deep generation has attracted increasing interest recently. Due to the lack of 3D face structures, most existing works~\cite{zakharov2019few,zhang2020freenet,ha2019marionette,wu2018reenactgan,kim2018deep,wiles2018x2face,burkov2020neural,fu2019high} require paired expression images or video sequences in training, which impairs their usability greatly. In addition, they usually condition on facial landmarks~\cite{zakharov2019few,zhang2020freenet,ha2019marionette,wu2018reenactgan,kim2018deep,fu2019high} that naturally entangle expressions, identity and poses rigidly. Several works~\cite{wiles2018x2face,burkov2020neural,siarohin2019first} attempt to use predicable latent features but they work on entangled expressions and poses. 

The proposed POCE requires only unpaired images in training but can edit facial expressions and head poses simultaneously. Besides, it converts face images into UV maps where facial expression editing and head pose generation can be achieved independently, which improves editing flexibility and controllability greatly.

\subsection{Image Completion} 
Image completion aims at filling missing pixels in images.
\cite{pathak2016context} presents Context Encoder to address hole-filling problems. 
\cite{yang2017high} introduces multi-scale neural patch synthesis for preserving contextual structures with high-frequency details. 
\cite{yu2019free} designs gated convolution to complete irregular holes. 
\cite{deng2018uv} proposes UV-GAN to complete self-occluded UV maps but requires large-scale incomplete/complete UV pairs (expensive and time-consuming to collect) in network training.
\cite{gecer2021ostec} introduces OSTeC for iterative texture completion, which needs to be optimized for each image in inference. 

The proposed UV completion network completes self-occluded UV maps in a self-supervised manner without requiring ground-truth UV map. In addition, the trained model can be applied to images collected from different people, which makes it accessible and scalable to different users and tasks.

\section{Proposed Method}

\subsection{Overview}
Fig.~\ref{fig:pipeline} shows the POCE pipeline. Inspired by the idea that UV maps project 3D texture data to a 2D pose-invariant template, we convert a face image into a UV map where expression editing and head pose generation can be disentangled elegantly. Given a \textit{Source} image, we first fit a 3D face model to sample a facial UV map which often contains certain missing regions due to self-occlusions. The \textit{Sampled UV} is then fed to the proposed \textit{UV Completion} which generates missing texture to produce a \textit{Completed UV}. The \textit{Completed UV} is further fed to the proposed \textit{UV Editing} that generates \textit{Edited UV} with target expression conditioned on \textit{Relative AUs} or transplanted from \textit{Reference UV}. With the head \textit{Pose Parameters}, the completed and edited UV map is finally rendered to a target head pose to achieve pose-controllable expression editing. More details about the \textit{UV Completion}, \textit{UV Editing}, \textit{3D Face Fitting} and \textit{Rendering} will be described in the ensuing subsections.

\subsection{UV Completion}
Due to self-occlusions, the facial UV map sampled from face image is usually incomplete with missing texture, which affects both realistic expression editing and new poses generation. We design a UV completion technique that learns to complete UV map in a patch-based manner, which can generate complete UV maps from incomplete ones and provide realistic texture for face rendering under various new poses.

\begin{figure}[t]
\begin{center}
\includegraphics[width=1.\linewidth]{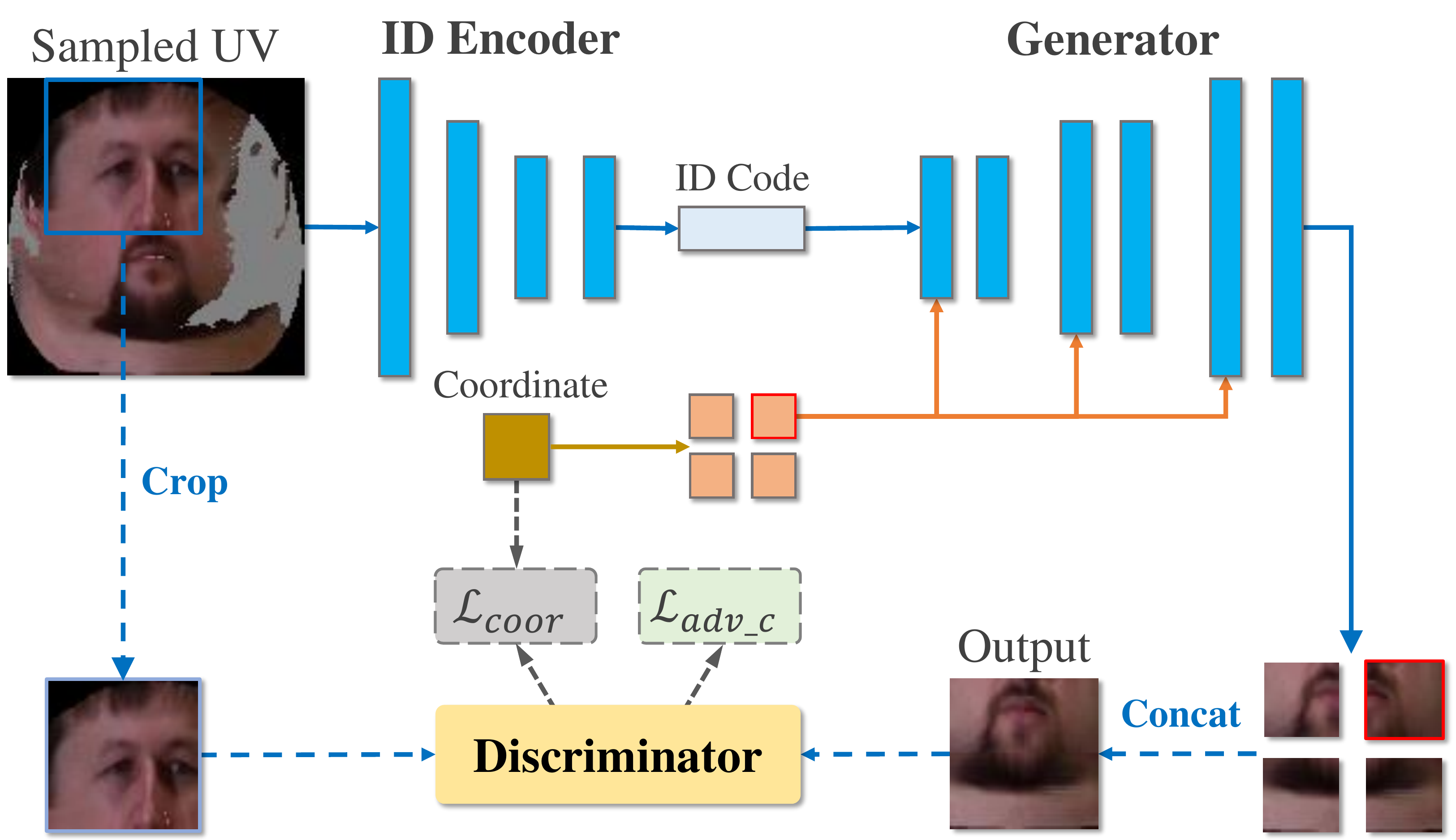}
\end{center}
\caption{
    Illustration of the proposed patch-based UV completion network: Given a \textit{Sampled UV} as input, an \textit{ID Encoder} first extracts the face identity information. A \textit{Generator} then takes the extracted \textit{ID Code} and the randomly sampled macro \textit{Coordinate} as input to generate a complete macro UV patch. Specifically, the \textit{Generator} first generates consecutive micro patches with the automatically-derived micro coordinates and stitches them to obtain the output macro patch. The real and generated macro UV patches are then forwarded to the \textit{Discriminator} for adversarial learning.
}
\label{fig:UV_completion}
\end{figure}

\subsubsection{Network Overview} 
As illustrated in Fig.~\ref{fig:UV_completion}, our patch-based UV completion network consists of an identity encoder ${E}_{I}$, a generator ${G}_{c}$ and a discriminator ${D}_{c}$. Given a sampled UV map, ${E}_{I}$ first extracts identity information which is shared among all patches of the same face image. ${G}_{c}$ then generates a complete macro UV patch based on the extracted identity information and a randomly sampled macro coordinate. Specifically, ${G}_{c}$ first generates consecutive micro patches with the micro coordinates derived from the sampled macro coordinate and stitches them to obtain the macro patch. ${D}_{c}$ aims to distinguish the generated macro patches against real ones (that cropped from incomplete UV maps) and it guides the generator to synthesize coherent contents for adjacent micro patches in an adversarial manner.
Once the model is trained, it synthesizes patches at all coordinates and stitches them as a completed UV map.

\subsubsection{Patch-based UV Completion} 
Most existing UV completion methods are trained with paired UV including a sampled incomplete UV map and a complete UV map (as the ground truth in supervised training) of the same person~\cite{deng2018uv}, but capturing complete face UV has various restrictions in equipment (requiring 3D scanner or multi-view cameras), portraiture light, etc. 
To the best of our knowledge, only one public dataset provides complete UV~\cite{deng2018uv}, but the UV maps are collected in lab environments, which do not generalize well to face images in the wild.
We design a patch-based UV completion network that can complete UV maps in a self-supervised manner without requiring complete ground-truth UV. Our design is inspired by the observations that UV maps are highly aligned (with universal per-pixel alignment) in the 2D pose-invariant template and the sampled UV is composed of visible texture in a continuous region.
We can thus crop complete UV patches from different locations of incomplete UVs though we do not have complete UV in training. 
With the cropped patches, our UV completion learns to complete UV in a patch-based manner conditioning on the spatial coordinates and face identity as extracted from incomplete UV. 

\begin{figure}[t]
\begin{center}
\includegraphics[width=1.\linewidth]{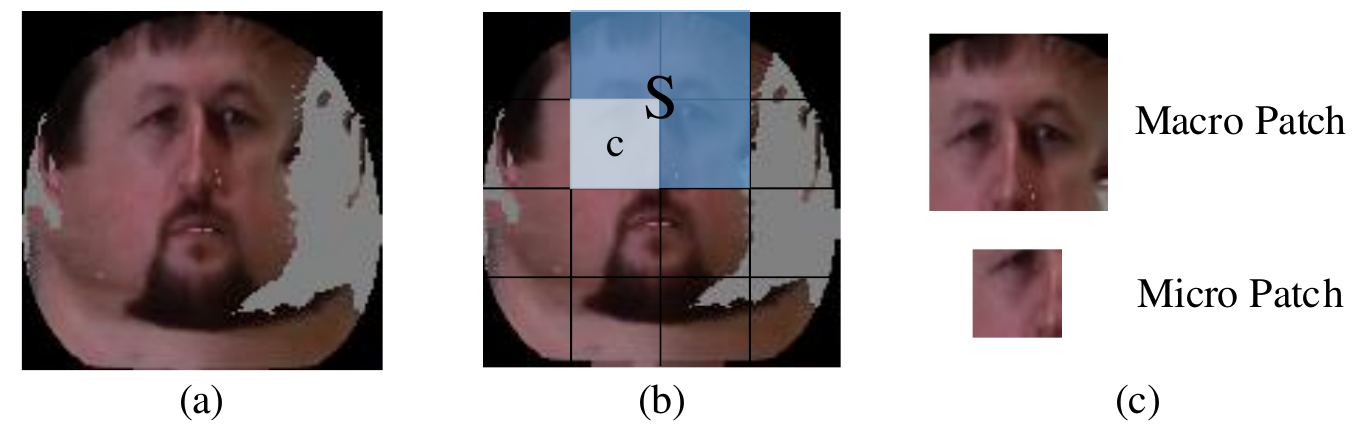}
\end{center}
\caption{
    Illustrations of the coordinate systems in the proposed UV completion: (a) shows a sampled UV, (b) illustrates the macro ($s$) and micro ($c$) coordinate systems used in our design, and (c) illustrates the macro and micro patches used for training the UV completion network.
}
\label{fig:coor_illustration}
\end{figure}

\subsubsection{Coordinate System} 
The proposed UV completion module learns to complete facial UV maps in a patch-based manner without requiring paired UV data for network training. The trained UV completion model can synthesize patches at all coordinates and stitch them to obtain a completed UV map. A coordinate system is thus required for providing spatial guidance. 
Our idea of designing the coordinate system is straightforward: it divides a UV map into multiple non-overlapping patches and assigns a unique coordinate to each patch to encode the spatial location information.  

However, directly generating patches and stitching them together often leads to artifacts around the patch boundaries and affects realistic UV completion.
Inspired by~\cite{lin2019coco}, we address this issue by training the network to generate consecutive micro patches and stitch them to form a macro patch, and introducing an adversarial loss to penalize the incoherent contents within the stitched macro patch. This strategy encourages the network to generate seamless boundaries for adjacent micro patches, which helps suppress the artifacts effectively. 
To this end, we design micro and macro coordinate systems for handling patches of different sizes as illustrated in Fig.~\ref{fig:coor_illustration}.

Specifically, the micro coordinate system divides a UV map into $m$ by $n$ micro patches without overlapping:
\begin{equation*}
C = 
\begin{pmatrix}
c_{1,1} & c_{1,2} & \cdots & c_{1,n} \\
c_{2,1} & c_{2,2} & \cdots & c_{2,n} \\
\vdots  & \vdots  & \ddots & \vdots  \\
c_{m,1} & c_{m,2} & \cdots & c_{m,n}
\end{pmatrix},
\end{equation*}
while the macro coordinate system is defined as:
\begin{equation*}
S = 
\begin{pmatrix}
s_{1,1} & s_{1,2} & \cdots & s_{1,n'} \\
s_{2,1} & s_{2,2} & \cdots & s_{2,n'} \\
\vdots  & \vdots  & \ddots & \vdots  \\
s_{m',1} & s_{m',2} & \cdots & s_{m',n'} 
\end{pmatrix},
\end{equation*}
where $s_{i,j} = [c_{i:i+p-1, j:j+q-1}]$, indicating that the macro patch is formed by $p \times q$ consecutive micro patches. $p$ and $q$ denote the number of micro patches that construct a macro patch in horizontal and vertical directions, respectively. With the given notations, we can easily derive $m' = m - p + 1$, $n' = n - q + 1$.  

We crop all macro UV patches from each incomplete UV in a raster-scan order according to the macro coordinate system, and empirically treat patches that contain more than 95$\%$ valid pixels as complete UV patches.

\begin{figure*}[t]
\begin{center}
\includegraphics[width=1.\linewidth]{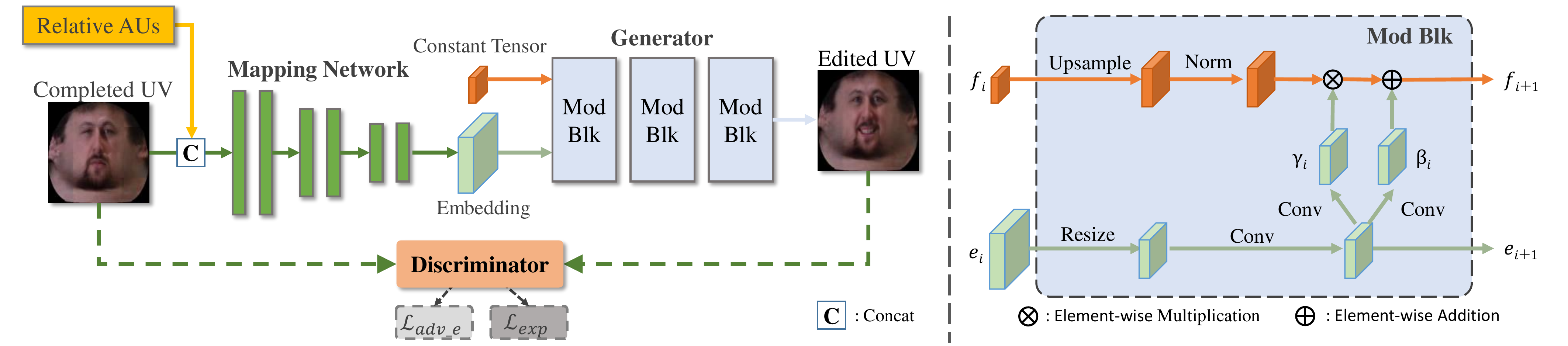}
\end{center}
\caption{
    Illustration of the proposed UV editing network: The \textit{Mapping Network} first transforms the \textit{Completed UV} and \textit{Relative AUs} to the latent \textit{Embedding} that encodes source identity attribute and desired facial expression.
    The \textit{Generator} then takes as input the \textit{Embedding} and the \textit{Constant Tensor} that encodes the coarse geometry prior shared among all UV maps to produce \textit{Edited UV}. 
    Detailed structure of the modulation block (\textit{Mod Blk}) is shown on the right.
}
\label{fig:UV_editing}
\end{figure*}

\subsubsection{Loss Functions}
The loss for training the UV completion network consists of four terms:
\begin{equation}
    \begin{aligned}
         \mathcal{L}_{C} = \mathcal{L}_{adv\_c} 
                                  + \lambda_{coor} \mathcal{L}_{coor} 
                                  + \lambda_{cont} \mathcal{L}_{cont} 
                                  + \lambda_{sym} \mathcal{L}_{sym},
        \label{formula:uvc_loss}
    \end{aligned}
\end{equation}
where the weights $\lambda$s are used to balance each component.

The first term is a patch-level adversarial loss that guides the discriminator to distinguish real/generated UV patches and the generator to generate complete and realistic patches: 
\begin{equation}
    \begin{aligned}
        \mathcal{L}_{adv\_c}  = 
                   & {\mathbb{E}} [D_{c} ( {\mathbf{T}^{i}_{G}})]
                    - {\mathbb{E}} [{D}_{c} ( {\mathbf{T}^{j}_{R}})] \\
                 & + \lambda_{c} {\mathbb{E}} 
                    [(\| \nabla {D}_{c} ({\mathbf{\hat{T}}}) \|_2 - 1)^2],
        \label{formula:uvc_D}
    \end{aligned}
\end{equation}
where ${\mathbf{T}^{i}_{G}}$ is the generated UV patch with coordinates $s^i$, ${\mathbf{T}^{j}_{R}}$ is the real complete UV patch with coordinates $s^j$ cropped from the visible region, $\mathbf{\hat{T}}$ is the interpolated data introduced in~\cite{gulrajani2017improved} and $\lambda_{c}$ is the balancing weight.

The second term is the coordinate regression loss that guides the generator to align the generated patches with the given spatial coordinates. Specifically, we add an auxiliary head on top of the discriminator to predict spatial coordinate, and apply L2 loss on the coordinates of both generated and real patches:
\begin{equation}
    \begin{aligned}
        \mathcal{L}_{coor} = \
          & {\mathbb{E}} [ \| D_{x} ( {\mathbf{T}^{i}_{G}}) - s^i \|_2^2]
            + {\mathbb{E}} [ \| D_{x} ( {\mathbf{T}^{j}_{R}}) - s^j \|_2^2] , 
        \label{formula:uvc_coor_loss}
    \end{aligned}
\end{equation}
where $D_{x}$ is the coordinate regression head on top of $D_{c}$.

The third term is the content loss that guides the generator to generate patches of the same identity. We penalize L1 difference between the generated patch and the one cropped from $ {UV_{sam}}$ at the same location (within visible region):
\begin{equation}
    \begin{aligned}
        \mathcal{L}_{cont}  = {\mathbb{E}} \| (  {\mathbf{T}^{i}_{G}} -  {\mathbf{T}^{i}_{R}} ) \otimes  {M^i} \|_1, 
        \label{formula:uvc_cont_loss}
    \end{aligned}
\end{equation}
where $\otimes$ denotes element-wise multiplication, $M^i$ is the visibility mask of the corresponding patch at $s^i$.

The last term is the symmetry loss that exploits facial symmetry to guide the network to generate missing texture:
\begin{equation}
    \begin{aligned}
        \mathcal{L}_{sym}  = \| ( {\mathbf{T}^{i}_{G}} - flip({\mathbf{T}^{-i}_{G}}) )\|_1, 
        \label{formula:uvc_sym_loss}
    \end{aligned}
\end{equation}
where $flip({\mathbf{T}^{-i}_{G}})$ is the horizontal flipped patch (at mirror location) of ${\mathbf{T}^{i}_{G}}$.

\subsection{UV Editing} 
\label{UV_editing}
With a completed UV map, the proposed UV editing network aims to edit its expression with minimal modification of facial identity features. 
We employ the widely adopted Facial Action Coding System (FACS)~\cite{friesen1978facial} to describe facial expressions in terms of the intensities of continuous Action Units (AUs).
Specifically, we exploit relative AUs as expression conditions for training the UV editing network and design a modulation-based generator that incorporates spatially-varying modulation to edit the facial expressions.
Once the model is trained, it allows to edit facial expressions that can be either controlled by the relative AUs that encode desired expression information or directly transplanted from a reference UV map via feature transfer.

\subsubsection{Network Overview} 
As illustrated in Fig.~\ref{fig:UV_editing}, our UV editing network consists of a mapping network $M_e$, a generator $G_e$ and a discriminator $D_e$. 
$M_e$ takes the completed UV map (by UV completion network) and relative AUs as input, and maps them to the latent embedding that encodes the identity attribute of source UV map as well as the desired facial expression.
$G_e$ then takes the embedding and a constant tensor that encodes the coarse geometry prior that is shared among all UV maps as input to generate an edited UV with target expression.
$D_e$ evaluates the photo-realism of edited UV and examines whether it contains desired expression information.

\subsubsection{Relative AUs} 
Inspired by~\cite{liu2019stgan,ling2020toward} that use difference vector to control image attributes, we propose to train the UV editing network conditioning on relative AUs, which has three benefits. 
First, relative AUs can better guide the network to focus on interested regions as compared with absolute AUs. Existing works~\cite{pumarola2018ganimation,wu2020cascade} feed absolute AUs and a source image to the network which first predicts an attention map to identify regions-of-interest and then performs editing. This requires the network to implicitly estimate source AUs and compare them with target AUs to generate the attention map. In contrast, utilizing relative AUs allows the residual information to be explicitly injected into the network, which guides the network to focus on the interested regions and makes it converge faster.
Second, using relative AUs helps generate more accurate editing especially when only specific facial regions require editing. To edit specific regions, models using absolute AUs need to estimate the corresponding AUs in source face, modify their intensities and generate edited expression with the modified AUs. However, the AU estimation may suffer from errors, which leads to editing of unrelated facial attributes. With relative AUs, models just need to modify relative AUs intensities of interested regions and set the intensities of remaining AUs to zero. This mitigates undesired AU manipulations and leads to more accurate editing.
Third, editing facial expressions conditioning on relative AUs facilitates the trained model to edit expressions by directly transplanting the expression from the reference UV map to the source UV via feature transfer (more details to be shared in Section~\ref{feature_transfer}).

Relative AUs ($AU_{rel}$) are defined as the difference between the AUs of the source image ($AU_{src}$) and target AUs ($AU_{tgt}$) that encode desired expressions:
\begin{equation}
    \begin{aligned}
        AU_{rel} = AU_{tgt} - AU_{src}.
        \label{formula:relative_AU}
    \end{aligned}
\end{equation}

\begin{figure}[t]
\begin{center}
\includegraphics[width=1.\linewidth]{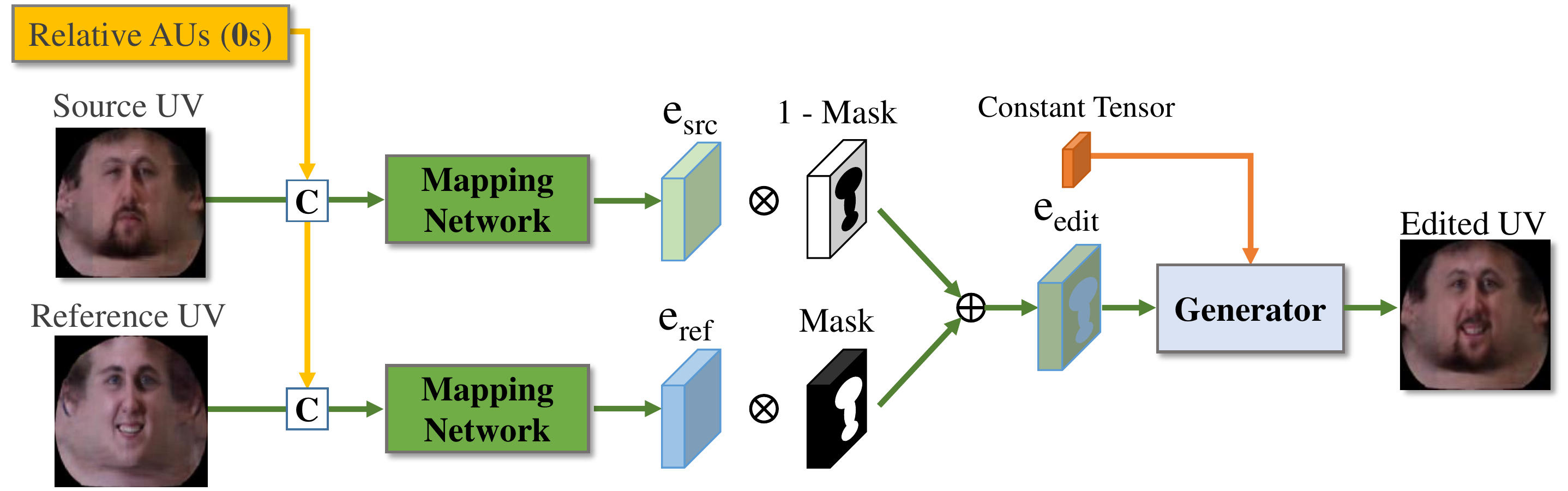}
\end{center}
\caption{
    Illustration of the proposed expression editing via feature transfer pipeline: The \textit{Source UV} and \textit{Reference UV} are first fed to the well-trained \textit{Mapping Network} together with the \textit{Zeroized Relative AUs} (where all elements are set to 0) to retrieve the corresponding embeddings ($e_{src}$ and $e_{ref}$) that encode the identity and expression of the original UV maps.
    $e_{src}$ is then blended with $e_{ref}$ with respective to a given \textit{Mask} to transfer the expression-related features from $e_{ref}$ to $e_{src}$.
    Finally, the edited embedding $e_{edit}$ is forwarded to the trained \textit{Generator} together with the \textit{Constant Tensor} to produce the \textit{Edited UV} that combines the facial expression of the reference UV and the identity attribute of the source UV.
}
\label{fig:UV_blending}
\end{figure}

\subsubsection{Modulation-based Generator}
\label{sec:UV_editing_generator}

Most existing facial expression editing models~\cite{pumarola2018ganimation,choi2018stargan,wu2020cascade} employ encoder-decoder architecture to generate the edited output. They first leverage an encoder to transform the source images to high-level representations, and then forward them to a decoder to produce the editing results. 
However, as discussed in~\cite{liu2019stgan}, the encoder-decoder architecture tends to discard fine details of source images, leading to blurry editing results.
In addition, the learned representations are unstructured which do not support facial expression editing via feature transfer (i.e., replacing partial representations of the source image with that of the reference image to achieve expression editing).

To mitigate the above issues, we design a modulation-based generator that incorporates spatially-varying modulation~\cite{park2019semantic,karras2020analyzing, kim2021stylemapgan} for expression editing in UV maps. 
The idea of designing the modulation-based generator is inspired by the observation that UV maps project 3D facial texture to a 2D pose-invariant template with universal per-pixel alignment, i.e., the facial features of different people (e.g., eyes, nose and mouth) are projected to similar location in the UV map. 
Hence, the UV maps of different people with different expressions share similar underlying geometry but vary in detailed texture information only. 
With this observation, the proposed modulation-based generator takes a constant tensor $t \in \mathbb{R}^{64 \times 8 \times 8}$ as initial input to encode the coarse geometry prior that is shared among all facial UV maps. 
It then forwards $t$ to multiple modulation blocks, which gradually produce feature maps of higher spatial resolution and inject detailed texture information into the feature maps (via spatially-varying modulation).
Finally, the generator transforms the output feature maps of the last modulation block to the edited UV that combines the desired facial expression and the identity attribute of the source UV. 
Since the generator is differentiable in its input, we follow~\cite{peebles2022gan} to optimize $t$ together with the network weights via backpropagation in network training. 

The modulation block modulates the feature maps with spatially-varying modulation parameters. 
These parameters are learned from the latent embedding (produced by a mapping network) that encodes the desired facial expression as well as the identity attribute of the source UV map.
The modulation operation of the $i$-th modulation block could be formulated as:
\begin{equation}
    \begin{aligned}
         f_{i+1} = (\gamma_{i} \otimes \frac{f_i - \mu_i}{\sigma_{i}}) \oplus \beta_{i},
        \label{formula:modulation}
    \end{aligned}
\end{equation}
where $f_i$ is the input feature maps ($f_0 = t$), $\mu_i$ and $\sigma_{i}$ are the mean and standard deviation of $f_i$, $\gamma_{i}$ and $\beta_{i}$ are the modulation parameters which have the same size as $f_i$, $\otimes$ and $\oplus$ denote the element-wise multiplication and addition operation, respectively.

With the modulation blocks, the designed generator can generate much sharper editing results than most existing methods~\cite{pumarola2018ganimation,choi2018stargan,wu2020cascade} that employ an encoder-decoder architecture.
In addition, the constant tensor and element-wise modulation explicitly help the generator build spatial connection between the modulated parameters and the output UV map. As a result, partial change in the modulated parameters will lead to local editing on the generated UV, which enables to edit facial expression via feature transfer.

\begin{figure*}[t]
\begin{center}
\includegraphics[width=1.\linewidth]{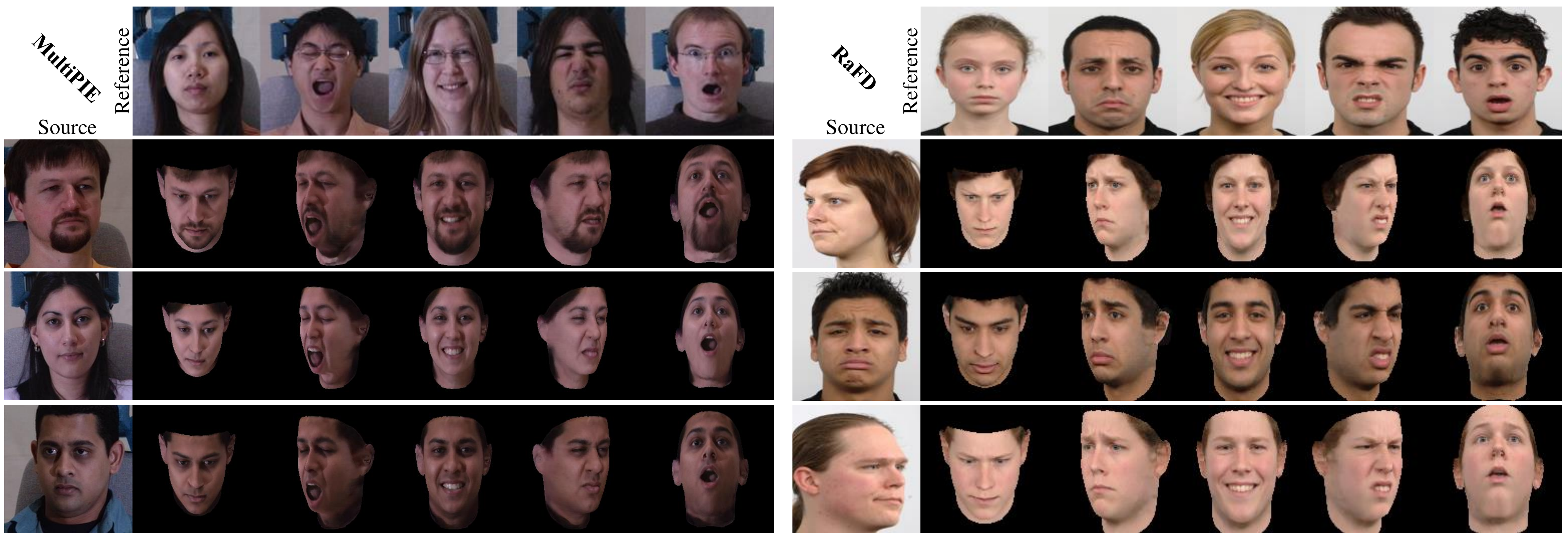}
\end{center}
\caption{
    Expression and pose editing by the proposed POCE over MultiPIE and RaFD: In each sub-figure, the first column shows the source images, and the first row shows the reference images with target expressions. The rest rows and columns show our editing. POCE is capable of editing facial expressions and head poses realistically and simultaneously in a disentangled manner.
}
\label{fig:pose_expr_combine}
\end{figure*}

\begin{figure*}[t]
\begin{center}
\includegraphics[width=1.\linewidth]{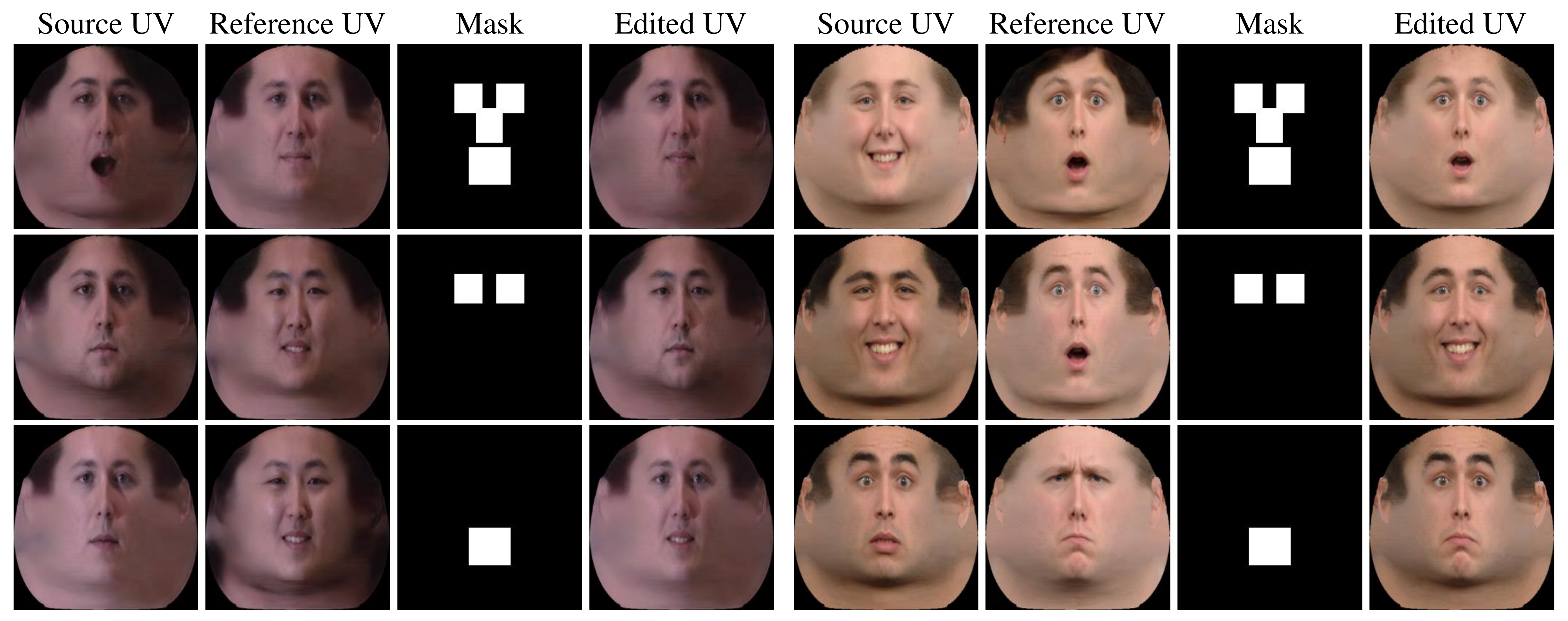}
\end{center}
\caption{
    Facial expression editing via feature transfer by POCE over MultiPIE and RaFD: Given a mask, POCE can transfer expression information around interested facial regions from reference UV to source UV and achieve editing of partial expression effectively. In each sub-figure, rows 1, 2 and 3 show the expression editing around eyes+nose+mouth, eyes only, and mouth only, respectively.
}
\label{fig:feature_blending}
\end{figure*}

\subsubsection{Expression Editing via Feature Transfer}
\label{feature_transfer}

In addition to edit expressions conditioning on relative AUs, the trained UV editing network allows to directly transplant expression from reference UV to source UV via feature transfer with respective to a given mask.
Note that the mask could be in any shape, which allows to simultaneously transfer expression around eyes, nose and mouth regions (that contain most expression-related information~\cite{althoff1999eye, hsiao2008two, wu2020cascade}), or just partial expression around interested facial region (e.g., eyes only or mouth only). 
In this way, we can flexibly transfer the desired expression from reference to source UV without requiring any expression annotation, which improves the editing flexibility greatly.

Fig.~\ref{fig:UV_blending} shows the expression editing via feature transfer pipeline. Given the source UV and the reference UV that provide identity and desired expression information, we first forward them to the trained mapping network together with the zeroized relative AUs (where all elements are set to 0) to retrieve the corresponding embeddings ($e_{src}$ and $e_{ref}$) that encode the identity and expression information of the original UV. We then apply alpha blending to $e_{src}$ and $e_{ref}$ to obtain the edited embedding $e_{edit}$:
\begin{equation}
    \begin{aligned}
         e_{edit} = m \otimes e_{ref} \oplus (1 - m) \otimes e_{src},
        \label{formula:feature_blending}
    \end{aligned}
\end{equation}
where $m$ is a binary mask that is resized by max pooling to match the spatial resolution of the embeddings, $\otimes$ and $\oplus$ are the same as Eq.~(\ref{formula:modulation}).
As discussed in Section~\ref{sec:UV_editing_generator}, the embedding is mapped to spatially-varying modulation parameters which inject detailed texture information into the feature maps in an element-wise manipulation manner. 
Transferring the reference embedding $e_{ref}$ around eyes, nose and mouth regions to that of $e_{src}$ thus transfers most expression information from the reference UV to the source UV. At the other end, partial embedding transfer leads to local expression editing, where the edited regions can be specified by masks as illustrated in Fig.~\ref{fig:feature_blending}.
Note we perform feature transfer on the modulation parameters which have larger spatial resolution (than the embedding) to achieve fine-grained manipulation in the experiments, but we explain the transfer operation on the embedding for simplicity.

Finally, we forward the edited embedding $e_{edit}$ as well as the constant tensor $t$ to the trained generator to produce the edited UV map that combines the identity attribute of source UV and the expression attribute of reference UV.

\subsubsection{Loss Functions}

The objective function for training the UV editing network consists of three terms:
\begin{equation}
    \begin{aligned}
         \mathcal{L}_{E} = \mathcal{L}_{adv\_e}
                                  + \lambda_{exp} \mathcal{L}_{exp} 
                                  + \lambda_{cyc} \mathcal{L}_{cyc}. 
        \label{formula:uve_loss}
    \end{aligned}
\end{equation}

The first term is an adversarial loss~\cite{gulrajani2017improved} for improving the photo-realism of the edited UV:
\begin{equation}
    \begin{aligned}
        \mathcal{L}_{adv\_e}  = 
                  &  {\mathbb{E}} [{D}_{e}( UV_{src})]
                    - {\mathbb{E}} [{D}_{e}( UV_{edit})]  \\
                  & + \lambda_{e} {\mathbb{E}} 
                    [(\| \nabla {D}_{e} ({\widehat{UV}}) \|_2 - 1)^2],
        \label{formula:uve_D}
    \end{aligned}
\end{equation}
\begin{equation}
    \begin{aligned}
        UV_{edit} = G_e(M_e(UV_{src}, AU_{rel}), t),
        \label{formula:UV_edit}
    \end{aligned}
\end{equation}
where $UV_{src}$ is the completed source UV map (by UV completion network), $UV_{edit}$ is the output of UV editing network, ${\widehat{UV}}$ is the interpolated data introduced in~\cite{gulrajani2017improved} and $\lambda_{e}$ is the balancing weight, respectively.

The second term is a conditional expression loss that guides the generator to generate a UV map with desired expression:
\begin{equation}
    \begin{aligned}
        \mathcal{L}_{exp} = 
                  & {\mathbb{E}} [ \| D_{y}( UV_{src} ) - AU_{src} \|_2^2] \\
                  & + {\mathbb{E}} [ \| D_{y}( UV_{edit} ) - AU_{tgt} \|_2^2],
        \label{formula:uve_cond_loss}
    \end{aligned}
\end{equation}
where $D_{y}$ is the AUs regression head on top of $D_{e}$.

The third term is a cycle reconstruction loss that guides the generator to keep facial identity and personal attributes of the source UV after editing:
\begin{equation}
    \begin{aligned}
         \mathcal{L}_{cyc}  = 
                  & {\mathbb{E}} \| UV_{src} -  G_e(M_e(UV_{src}, 0), t) \|_1 \ + \\
                  & {\mathbb{E}} \| UV_{src} - G_e(M_e(UV_{edit}, -AU_{rel}), t) \|_1.
        \label{formula:uve_cyc_loss}
    \end{aligned}
\end{equation}

\begin{figure*}[t]
\begin{center}
\includegraphics[width=1.\linewidth]{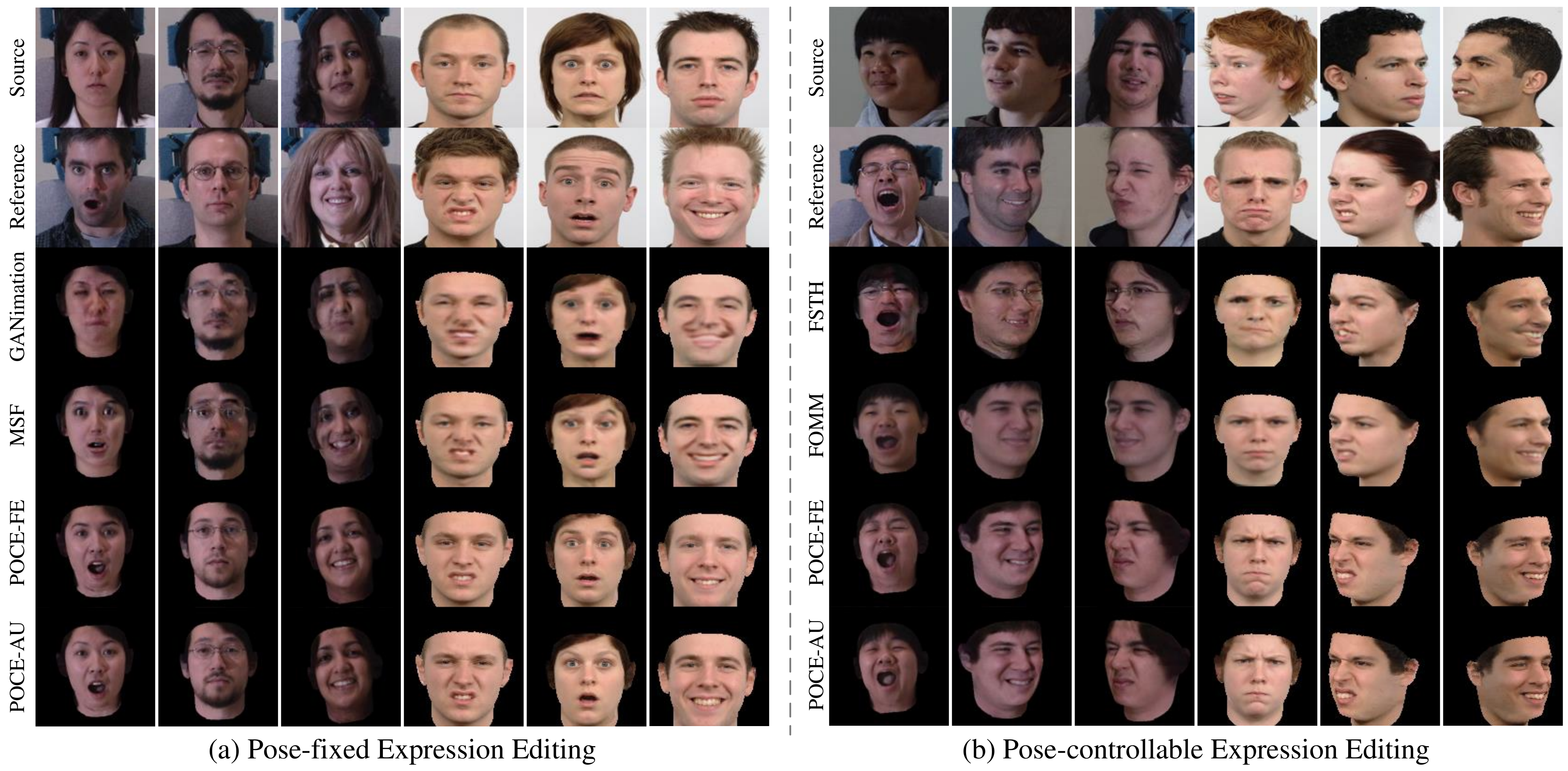}
\end{center}
\caption{
    Expression editing by POCE and the state-of-the-art methods over MultiPIE and RaFD: In each sub-figure, columns 1-3 show the editing over MultiPIE images while columns 4-6 show the editing over RaFD images. 
    POCE-FE and POCE-AU indicate the POCE editing results generated via feature transfer and conditioning on relative AUs, respectively.
    It can be seen that POCE can produce more realistic editing with better details and less artifacts than the state-of-the-art methods in both pose-fixed and pose-controllable expression editing.
}
\label{fig:qualitative_comparison_pose_combine}
\end{figure*}

\subsection{3D Face Fitting and Rendering}

\subsubsection{3D Face Fitting}
We fit a pre-trained 3DMM to face images to derive facial UV maps. 
A number of open-source 3DMM models are available~\cite{zhu2016face,feng2018joint,bulat2017far} and we adopt 3DDFA~\cite{zhu2016face}. 
Specifically, we apply 3DDFA to estimate 3D face shape $\mathcal{{S}}$ and head pose $\mathcal{{P}}$ from the face image $I$:
\begin{equation}
    \begin{aligned}
        \{ \mathcal{{S, P}} \} = Fitting(I),
        \label{formula:fitting}
    \end{aligned}
\end{equation}
where $\mathcal{{P}}$ is parameterized by Euler’s angles (\textit{pitch}, \textit{yaw}, \textit{roll}), translation and a face scale factor.
The 3D face is then projected to the image plane with weak perspective projection and the facial UV map ${UV_{sam}}$ can be sampled from $I$ with the standard rasterization pipeline~\cite{zhou2020rotate}:
\begin{equation}
    \begin{aligned}
          \{{UV_{sam}, M\}} = \mathbf{F} (I, \mathcal{{V(S,P)}}),
        \label{formula:sampling}
    \end{aligned}
\end{equation}
where $\mathbf{F}$ denotes the sampling operation, $\mathcal{{V}}$ denotes the projection operation and ${M}$ is a binary visibility mask of the sampled UV.

\subsubsection{Rendering}
The sampled UV map is fed to our UV completion and UV editing modules for generating complete texture and editing expression, respectively. 
With a completed and edited UV map $UV_{edit}$, we can render it to target pose $\mathcal{{P}}_{tgt}$ (which could be either tuned by user or estimated from a reference image) with an off-the-shelf renderer to achieve pose-controllable expression editing:
\begin{equation}
    \begin{aligned}
          {I_{output}} = \mathbf{R} (UV_{edit}, \mathcal{{S, P}}_{tgt}),
        \label{formula:render}
    \end{aligned}
\end{equation}
where $\mathbf{R}$ denotes the renderer. In our implementation, we use an open-sourced renderer~\cite{kato2018neural} to perform rendering without any training.

\renewcommand\arraystretch{1.2}
\begin{table*}[t]
\small 
\renewcommand\tabcolsep{3pt}
\caption{Quantitative comparisons of pose-fixed expression editing between POCE and SOTA methods over MultiPIE and RaFD.}
\centering
\begin{tabular}{|m{1.9cm}<{\centering}|
                m{2.4cm}<{\centering}|
                m{1.0cm}<{\centering}
                m{1.0cm}<{\centering}
                m{1.0cm}<{\centering}
                m{0.9cm}<{\centering}
                m{1.1cm}<{\centering}
                m{1.1cm}<{\centering}|} \hline
Dataset & Methods & IED $\downarrow$ & EXD $\downarrow$ & FID $\downarrow$ & IS $\uparrow$ & RA $\uparrow$ & ES $\uparrow$ 
\\\hline

\multirow{5}{*}{MultiPIE~\cite{gross2010multi}} 
& \color{gray}{Real}  & \color{gray}{-} & \color{gray}{-} & \color{gray}{0.00} & \color{gray}{1.578} & \color{gray}{4.66} & \color{gray}{-} \\
& GANimation~\cite{pumarola2018ganimation} & 0.470
& 0.433 & 31.76 & 1.405 & 1.72 & 1.68 \\
& MSF~\cite{ling2020toward} & 0.406 & 0.408 & 25.04 & 1.434 & 2.95 & 3.13 \\
& POCE (FE) & 0.365 & \textbf{0.342} & 20.82 & 1.510 & 4.04 & \textbf{4.47} \\
& POCE (AU) & \textbf{0.348} & 0.357 & \textbf{20.43} & \textbf{1.522} & \textbf{4.13} & 4.32 
\\\hline

\multirow{5}{*}{RaFD~\cite{langner2010presentation}} 
& \color{gray}{Real} & \color{gray}{-} & \color{gray}{-} & \color{gray}{0.00} & \color{gray}{1.819} & \color{gray}{4.38} & \color{gray}{-}\\
& GANimation~\cite{pumarola2018ganimation} & 0.632 & 0.329 & 14.18 & 1.586 & 1.46 & 2.37\\
& MSF~\cite{ling2020toward} & 0.493 & 0.306 & 9.55 & 1.644 & 3.20 & 3.41\\
& POCE (FE) & 0.434 & \textbf{0.274} & 7.53 & 1.726 & \textbf{3.91} & \textbf{4.52} \\
& POCE (AU) & \textbf{0.427} & 0.281 & \textbf{7.41} & \textbf{1.737} & 3.88 & 4.33
\\\hline

\end{tabular}
\label{tab_pose1}
\end{table*}

\renewcommand\arraystretch{1.2}
\begin{table*}[t!]
\small 
\renewcommand\tabcolsep{3pt}
\caption{Quantitative comparisons of pose-controllable expression editing between POCE and SOTA methods over MultiPIE and RaFD.}
\centering
\begin{tabular}{|m{1.9cm}<{\centering}|
                m{2.4cm}<{\centering}|
                m{1.0cm}<{\centering}
                m{1.0cm}<{\centering}
                m{1.0cm}<{\centering}
                m{0.9cm}<{\centering}
                m{1.1cm}<{\centering}
                m{1.1cm}<{\centering}|} \hline
Dataset & Methods & IED $\downarrow$ & EXD $\downarrow$ & FID $\downarrow$ & IS $\uparrow$ & RA $\uparrow$ & ES $\uparrow$ 
\\\hline

\multirow{5}{*}{MultiPIE~\cite{gross2010multi}} 
& \color{gray}{Real} & \color{gray}{-} & \color{gray}{-} & \color{gray}{0.00} & \color{gray}{1.681} & \color{gray}{4.41} & \color{gray}{-} \\
& FSTH~\cite{zakharov2019few} & 0.872 & 0.518 & 27.04 & 1.499 & 1.65 & 2.07 \\
& FOMM~\cite{siarohin2019first} & 0.858 & 0.543 & 24.57 & 1.587 & 3.08 & 3.38 \\
& POCE-FE & 0.831 & \textbf{0.479} & 20.01 & 1.635 & 3.93 & \textbf{4.25} \\
& POCE-AU & \textbf{0.820} & 0.492 & \textbf{19.46} & \textbf{1.640} & \textbf{4.01} & 4.16
\\\hline

\multirow{5}{*}{RaFD~\cite{langner2010presentation}} 
& \color{gray}{Real} & \color{gray}{-} & \color{gray}{-} & \color{gray}{0.00} & \color{gray}{1.767} & \color{gray}{4.29} & \color{gray}{-} \\
& FSTH~\cite{zakharov2019few} & 1.174 & 0.409 &  13.59 & 1.610 & 1.81 & 1.49 \\
& FOMM~\cite{siarohin2019first} & 1.085 & 0.424 &  12.28 & 1.626 & 2.87 & 2.61\\
& POCE-FE & 0.924 & \textbf{0.338} & 10.73 & 1.652 & 3.50 & 3.79 \\
& POCE-AU & \textbf{0.897} & 0.366 & \textbf{10.35} & \textbf{1.663} & \textbf{3.62} & \textbf{3.85}
\\\hline

\end{tabular}
\label{tab_pose2}
\end{table*}

\section{Experiments}

\subsection{Settings}

\noindent{\bf Datasets: }
The proposed POCE is evaluated over datasets MultiPIE~\cite{gross2010multi} and Radboud Faces (RaFD)~\cite{langner2010presentation}. MultiPIE consists of more than 750,000 images of 337 identities showing different facial expressions.
RaFD contains 8,040 facial expression images of 67 participants collected from different viewpoints. 
We randomly sample 90$\%$ images for training and the rest for testing for both datasets.

\smallskip
\noindent{\bf Evaluation Metrics: }
We perform quantitative evaluations and comparisons with several widely adopted metrics as follows: 

\noindent $\bullet$ Identity Embedding Distance (IED). IED measures L2 distances between the embedded features of source and edited faces that are extracted by a pre-trained face recognition model~\cite{huang2020curricularface}. A lower IED indicates better identity preservation of the edited face images. 

\noindent $\bullet$ Expression Distance (EXD). EXD measures L2 distances between the AUs intensities of reference and edited faces estimated by OpenFace~\cite{baltrusaitis2018openface}. Lower EXDs mean higher similarity between the expressions of the edited and reference faces. 

\noindent $\bullet$ Fr\'{e}chet Inception Distance (FID)~\cite{heusel2017gans} and Inception Score (IS)~\cite{salimans2016improved}. FID and IS are computed based on the extracted features of pre-trained models~\cite{szegedy2016rethinking,szegedy2017inception}. Lower FID and higher IS indicate better image quality of the edited faces.

\noindent $\bullet$ Subjective evaluations. 
We also conducted Amazon-Mechanical-Turk (AMT) user studies to evaluate the perceptual realism of the edited images. 
Specifically, the subjects are tasked to give their ratings (1-5) to evaluate the editing quality based on two criteria.
The first is Realism Assessment (RA), where the subjects are presented with real and edited images and tasked to tell whether the images are real or fake. 
The second is Expression Similarity (ES), where the subjects are presented with a pair of images that consist of a reference image and an edited image (by different methods) and tasked to evaluate whether they contain similar facial expression.
We report mean opinion score of the ratings collected from the AMT users, where larger scores indicates better perceptual quality of the edited images.

\noindent{\bf Implementation Details: }
The UV completion and UV editing networks are trained separately.
Specifically, we first train the UV completion network with the sampled UV until it converges. Then we use the completed UV maps (generated by the well-trained UV completion network) to train the UV editing network.
The training is conducted on a single GeForce RTX 2080 Ti GPU with 11 GB memory and the size of UV maps is set to $256 \times 256$ in all the experiments.
 
\noindent $\bullet$ UV Completion Training Details: We use Adam optimizer~\cite{kingma2014adam} with $\beta_{1}=0$ and $\beta_{2}=0.999$ to optimize the parameters. 
We set $\lambda_{c}$, $\lambda_{coor}$, $\lambda_{cont}$ and $\lambda_{sym}$ to 100, 10, 1 and 10 to balance different losses.
The size of the micro patch is set to $64 \times 64$ and $2 \times 2$ micro patches are stitched to construct a macro patch.
The batch size is 2 and the total number of epochs is 30.
The learning rate is 1e-4 for the first 15 epochs, then it linearly decays to 0 over another 15 epochs.

\noindent $\bullet$ UV Editing Training Details: We use Adam optimizer~\cite{kingma2014adam} with $\beta_{1}=0.5$ and $\beta_{2}=0.999$ to optimize the parameters.
We set $\lambda_{e}$, $\lambda_{exp}$ and $\lambda_{cont}$ to 10, 1 and 1 to balance different losses.
The batch size is set to 4. The total number of epochs is set to 100.
The learning rate is set to 2e-4 for the first 50 epochs, then it linearly decays to 0 over another 50 epochs.

\begin{figure*}[t]
\begin{center}
\includegraphics[width=1.\linewidth]{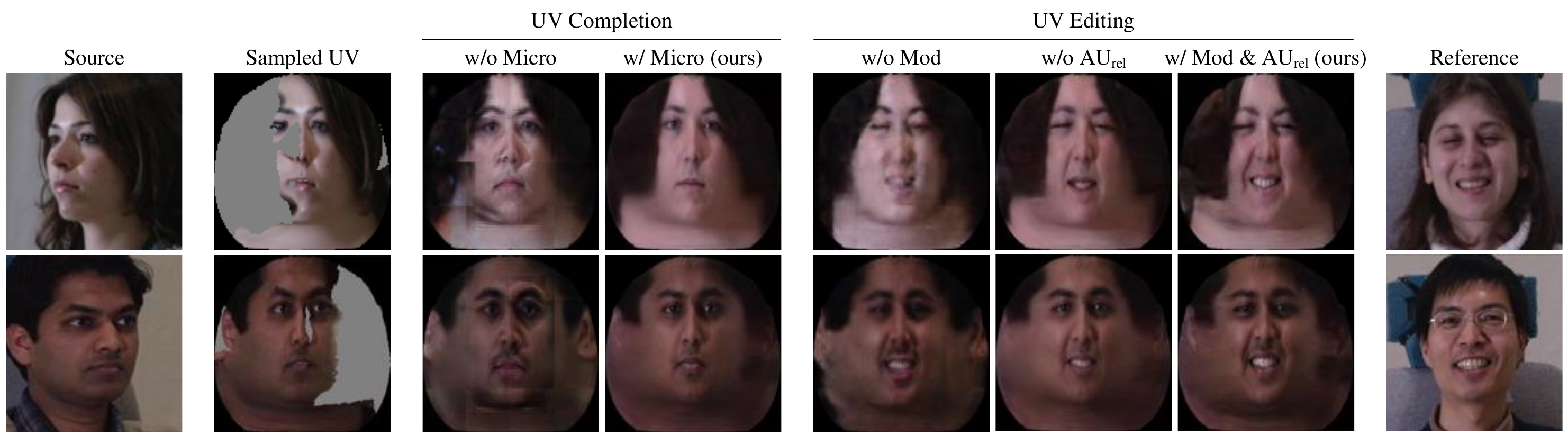}
\end{center}
\caption{
    Qualitative ablation study of POCE on MultiPIE: The proposed micro patch design helps generate better details and less artifacts as compared with \textit{w/o Micro} in UV completion. 
    In addition, including the modulation block (\textit{Mod}) and relative AUs (\textit{$AU_{rel}$}) helps generate more realistic facial details and more consistent expression intensity in UV editing. 
}
\label{fig:ablation}
\end{figure*}

\begin{figure}[t]
\begin{center}
\includegraphics[width=1.\linewidth]{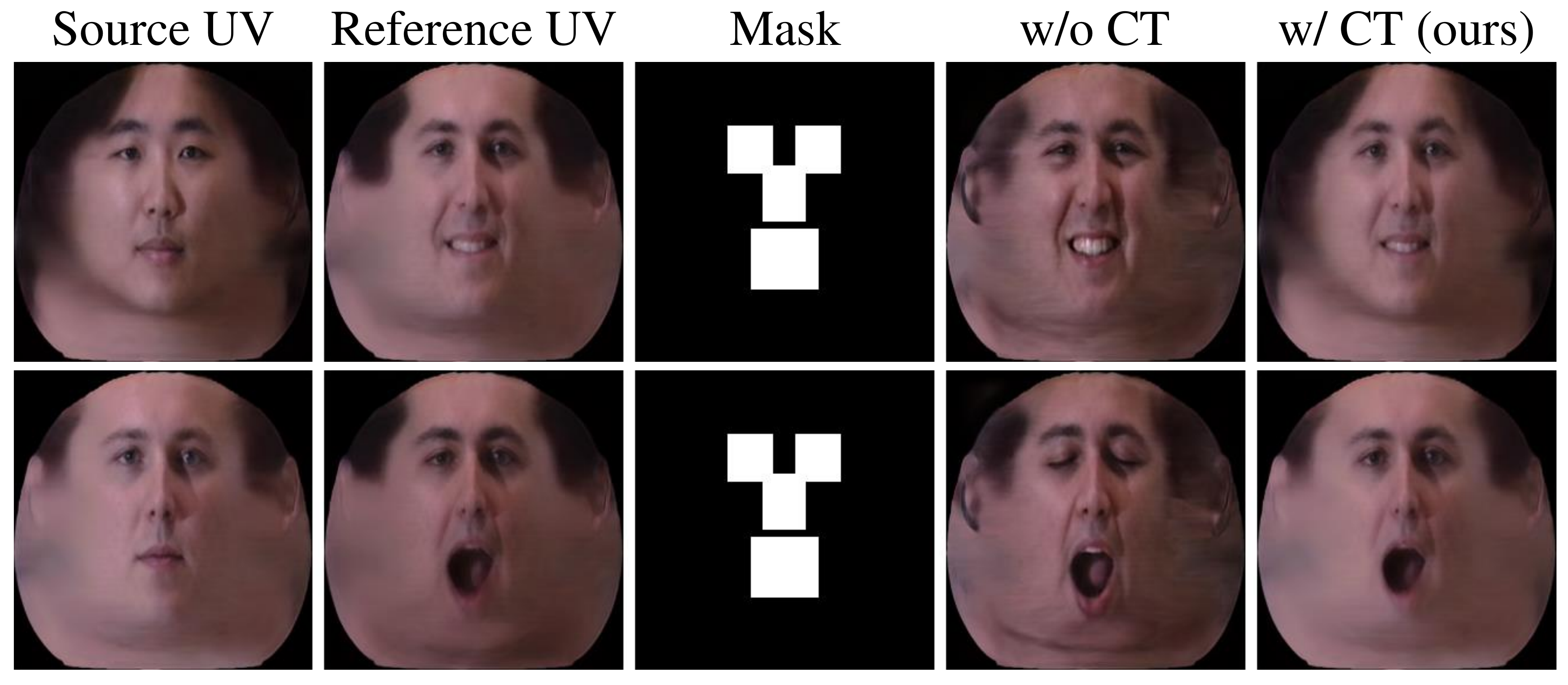}
\end{center}
\caption{
    Qualitative ablation study of POCE's expression editing via feature transfer on MultiPIE: Without using the constant tensor (CT), the produced UV maps are degraded by inconsistent expression intensity and altered person identity after feature transfer.
}
\label{fig:ablation_constant_tensor}
\end{figure}

\subsection{Experimental Results}

\noindent\textbf{Pose-Controllable Expression Editing: } 
The proposed POCE can edit facial expressions and head poses simultaneously and it just requires unpaired images in training. Fig.~\ref{fig:pose_expr_combine} illustrates the editing of a few sample images from MultiPIE~\cite{gross2010multi} and RaFD~\cite{langner2010presentation}. 
Specifically, we first use OpenFace~\cite{baltrusaitis2018openface} to extract the AUs intensities of source and reference images to derive relative AUs, then feed the source images as well as the relative AUs to POCE to produce the editing results.
It can be seen that POCE can edit expressions and poses simultaneously without changing the facial identity. More importantly, POCE edits expressions and poses in a disentangled manner, which translates to great editing flexibility by allowing to edit expressions only, poses only, or both of them with expressions and poses from different reference images. The superior editing usability and flexibility are largely attributed to the proposed disentangling approach within the UV space.

\smallskip
\noindent\textbf{Facial Expression Editing via Feature Transfer: } 
Beyond expression editing over relative AUs, POCE allows direct expression transfer (from reference UV to source UV) via feature transfer with given masks.
Fig.~\ref{fig:feature_blending} illustrates the editing of a few samples from MultiPIE~\cite{gross2010multi} and RaFD~\cite{langner2010presentation}.
Specifically, the trained UV completion network first generates the completed source and reference UV maps, which are then fed to the UV editing network (together with masks that indicate the regions to be edited) to produce the edited UV.
Note that the mask could be in any shape, which allows to flexibly transfer expression information around interested facial regions from reference UV to source UV without any expression annotations.
The superior editing flexibility is largely attributed to our proposed modulation-based generator.

\renewcommand\arraystretch{1.2}
\begin{table}[t!]
\small 
\renewcommand\tabcolsep{3pt}
\caption{Quantitative ablation study of POCE on MultiPIE.}
\centering
\begin{tabular}{|m{2.5cm}<{\centering}||
                m{1.1cm}<{\centering}
                m{1.1cm}<{\centering}
                m{1.1cm}<{\centering}
                m{1.1cm}<{\centering}|} \hline
\textbf{Models} & IED $\downarrow$ & EXD $\downarrow$ & FID $\downarrow$ & IS $\uparrow$ 
\\\hline
w/o Micro & 0.872 & 0.535 & 25.19 & 1.581  \\
w/o Mod & 0.866 & 0.547 & 26.34 & 1.560 \\
w/o AU$_{rel}$ & 0.828 & 0.513 & 19.78 & 1.631  \\
w/o CT & 0.839 & 0.506 & 20.12 & 1.625  \\
\hline
POCE & \textbf{0.820} & \textbf{0.492} & \textbf{19.46} & \textbf{1.640}
  \\\hline
\end{tabular}
\label{table:ablation_evaluation}
\end{table}

\smallskip
\noindent\textbf{Qualitative Evaluation: }
We first compare POCE with the state-of-the-art pose-fixed expression editing methods GANimation~\cite{pumarola2018ganimation} and MSF~\cite{ling2020toward} in Fig.~\ref{fig:qualitative_comparison_pose_combine}(a), where non-facial regions are masked for better comparisons. All methods are trained with continuous AUs intensities extracted by OpenFace~\cite{baltrusaitis2018openface}.
In particular, POCE-FE and POCE-AU indicate the editing results generated via feature transfer and conditioning on relative AUs, respectively.
As Fig.~\ref{fig:qualitative_comparison_pose_combine}(a) shows, GANimation~\cite{pumarola2018ganimation} and MSF~\cite{ling2020toward} tend to generate blurs and artifacts and even corrupted facial regions around mouths. 
POCE can instead generate more realistic expressions with much less blurs and artifacts, and the generated faces are also clearer and sharper.  
The better editing is largely attributed to our designed modulation-based generator that helps generate better facial details. 
In addition, GANimation~\cite{pumarola2018ganimation} and MSF~\cite{ling2020toward} cannot generate new poses in editing, while POCE can edit expressions and poses simultaneously as shown in Fig.~\ref{fig:pose_expr_combine}. 

We then compare POCE with FSTH~\cite{zakharov2019few} and FOMM~\cite{siarohin2019first} for pose-controllable expression editing in Fig.~\ref{fig:qualitative_comparison_pose_combine}(b), where non-facial regions are masked for better comparisons. 
FSTH~\cite{zakharov2019few} and FOMM~\cite{siarohin2019first} are trained with paired data, and they generate edited faces conditioned on facial landmarks and the predicted optical flows, respectively. 
As Fig.~\ref{fig:qualitative_comparison_pose_combine}(b) shows, FSTH~\cite{zakharov2019few} and FOMM~\cite{siarohin2019first} tend to generate degraded facial details and inconsistent expressions with reference images, and they fail to preserve facial identity in some editing (e.g., samples in column 3 and 5). One possible reason is landmarks and the predicted flows cannot capture fine-grained expression details and inevitably encode certain identity information of the reference face. POCE can instead generate more realistic expressions with less blurs and it preserves facial identity better as well.
In addition, FSTH~\cite{zakharov2019few} and FOMM~\cite{siarohin2019first} edit expressions and poses in an entangled manner, while POCE disentangles expressions and poses editing as shown in Fig.~\ref{fig:pose_expr_combine}.

\begin{figure*}[t!]
\begin{center}
\includegraphics[width=1.\linewidth]{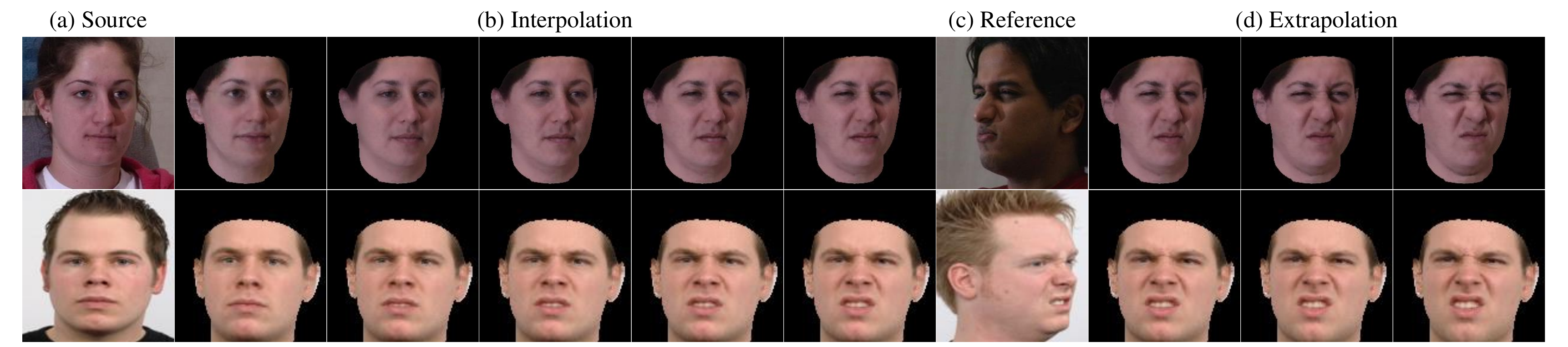}
\end{center}
\caption{
    Continuous expression editing by POCE: Given source images in (a) and reference images in (c), POCE can edit expressions by either interpolation or extrapolation over the relative AUs as shown in (b) and (d).
}
\label{fig:AU_smooth}
\end{figure*}

\begin{figure*}[t!]
\begin{center}
\includegraphics[width=1.\linewidth]{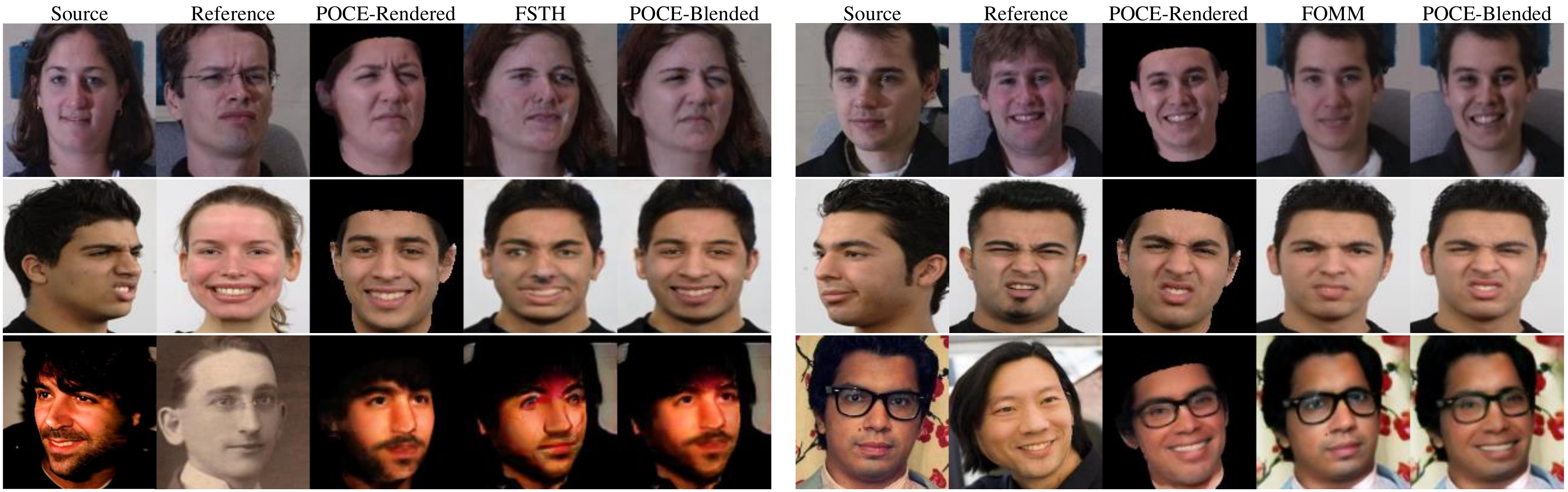}
\end{center}
\caption{
    Face blending by the proposed POCE over MultiPIE (first row), RaFD (second row) and wild images (last row): The POCE-rendered images can be seamlessly blended with FSTH~\cite{zakharov2019few} and FOMM~\cite{siarohin2019first} outputs to generate realistic images.
}
\label{fig:face_blending}
\end{figure*}

\smallskip
\noindent\textbf{Quantitative Evaluation: }
Table~\ref{tab_pose1} shows quantitative comparisons of pose-fixed expression editing over MultiPIE~\cite{gross2010multi} and RaFD~\cite{langner2010presentation}. We can see POCE consistently performs the best over all metrics. 
For identity preservation, POCE-AU achieved the best IED compared with the state-of-the-art methods.
For expression editing accuracy, POCE-AU obtained the best EXD as it directly blends the expression-related features from reference to source UV, which leads to more accurate expression editing.
For perceptual quality, POCE-AU consistently outperform other competitive methods on MultiPIE~\cite{gross2010multi} and RaFD~\cite{langner2010presentation}, with an improvement of FID by 4.61 and 2.14 as well as IS by 6.14\% and 5.66\%. 
Further, the POCE edited images obtained clearly higher scores in both user studies, which demonstrates the superiority of POCE in generating more realistic editing and more consistent expressions with respect to the reference images.

We also compare POCE with FSTH~\cite{zakharov2019few} and FOMM~\cite{siarohin2019first} over MultiPIE~\cite{gross2010multi} and RaFD~\cite{langner2010presentation} with the same metrics. 
As Table~\ref{tab_pose2} shows, POCE clearly outperforms FSTH~\cite{zakharov2019few} and FOMM~\cite{siarohin2019first} under different metrics, which suggests POCE can achieve more realistic synthesis in pose-controllable expression editing.

\subsection{Discussion}

\noindent\textbf{Ablation Study: } 
We first study how our proposed micro patch generation, modulation block and relative AUs contribute to the UV completion and UV editing.
As Fig.~\ref{fig:ablation} shows, including the proposed micro patch generation improves the UV completion in both smoothness and realism as compared with direct UV completion (w/o Micro) which produces inconsistent facial texture. 
In addition, the edited UV maps suffer from blurred facial details if we replace the modulation block with transposed convolutional layer (w/o Mod), and they tend to contain inconsistent expression intensity with the reference images (e.g., the mouth region) without using relative AUs (w/o AU$_{rel}$). 
Including the modulation block and relative AUs clearly produces sharper details and better expression consistency with the reference images.

We next study the contribution of the proposed constant tensor (CT) in expression editing via feature transfer.
As described in Section~\ref{sec:UV_editing_generator}, the constant tensor encodes coarse geometry prior that is shared among all UV maps. The generator takes the same constant tensor as initial input, gradually injects detailed texture information into it conditioning on the modulation parameters, and finally produces edited UV with desired expressions. 
In this process, the constant tensor serves as an anchor that helps the generator build spatial connection between the modulated parameters and the output UV map, which enables expression editing via feature transfer.
The generator fails to learn such connection if we replace the constant tensor with randomly sampled Gaussian noise in network training (w/o CT), leading to inconsistent expression intensity and altered identity information in the produced UV maps as illustrated in Fig.~\ref{fig:ablation_constant_tensor}.

We also conduct quantitative experiments to evaluate the contributions of each component. Table~\ref{table:ablation_evaluation} shows the experimental results. The quantitative experimental results further verify the effectiveness of our proposed techniques.

\noindent\textbf{Continuous Expression Editing: } 
POCE can be easily adapted to generate continuous facial expressions. Given relative AUs between the source and the reference images, intermediate relative AUs of different stages can be simply derived by linear interpolation. Continuous facial expressions can thus be generated from the source images and the interpolated AUs by POCE as illustrated in Fig.~\ref{fig:AU_smooth}.

\noindent\textbf{Face Blending: } 
As facial UV maps are sampled from ear-to-ear facial region, the POCE rendered images do not capture hair and image background. We address this issue by introducing face blending. Specifically, we first generate the target head pose with existing pose-controllable expression editing model, which usually contains corrupted facial features and inconsistent expression with the reference image, and then blend it with the POCE-rendered expression via Poisson editing~\cite{perez2003poisson}. In our experiment, we blend the POCE-rendered faces to FSTH~\cite{zakharov2019few} and FOMM~\cite{siarohin2019first}. As Fig.~\ref{fig:face_blending} shows, the POCE-rendered images can be seamlessly blended to FSTH and FOMM outputs to generate realistic images.

\noindent\textbf{Expression Editing on Wild Images: } 
POCE can be adapted to handle wild images as shown in the last row in Fig.~\ref{fig:face_blending}, where the model is jointly trained on MultiPIE~\cite{gross2010multi}, RaFD~\cite{langner2010presentation} and 300W-LP datasets~\cite{zhu2016face}.
As Fig.~\ref{fig:face_blending} shows, POCE can edit facial expressions and head poses successfully while maintaining identity information and personal attributes well (e.g., mustache and eyeglasses).

\noindent \textbf{Ethical Considerations: } 
With the convenience of generating pose-controllable expression editing faces from unpaired images, POCE could be misused by immoralists to spread misinformation.
To avoid improper uses, we will include watermark to generated faces to indicate that they are synthetic.

\section{Conclusion}

This paper presents POCE for pose-controllable expression editing with just unpaired training data.
Our method converts face images into UV maps and disentangles facial expression editing and head pose generation elegantly. 
We propose self-supervised UV completion and weakly-supervised UV editing to complete the missing facial textures on the sampled UV maps and edit expression information on the completed UV, respectively. 
Extensive experiments show that POCE can generate realistic facial expressions and head poses simultaneously.
We expect that POCE will inspire new insights and attract more interests for better facial expression editing in the near future.

\section{Acknowledgments}
This work is funded by the Ministry of Education, Singapore, under the Tier-1 Project RG94/20.

\bibliographystyle{IEEEtran}
\bibliography{egbib}

\vfill

\end{document}